\pdfoutput=1

\documentclass[11pt]{article}

\usepackage[final]{acl}

\usepackage{times}
\usepackage{latexsym}
\usepackage{hyperref}
\usepackage[T1]{fontenc}

\usepackage[utf8]{inputenc}

\usepackage{microtype}

\usepackage{inconsolata}

\usepackage{graphicx}

\title{Instructions for *ACL Proceedings}

\usepackage{times}
\usepackage{latexsym}

\usepackage[T1]{fontenc}

\usepackage[utf8]{inputenc}

\usepackage{microtype}

\usepackage{inconsolata}

\usepackage{graphicx}

\usepackage{times}
\usepackage{latexsym}

\usepackage{booktabs}
\usepackage{siunitx}
\usepackage{amsmath,amssymb,amsthm}
\DeclareMathOperator*{\argmax}{argmax}
\usepackage{multirow}

\usepackage{amsmath}
\usepackage{tikz}
\usepackage{array}
\newcolumntype{C}[1]{>{\centering\arraybackslash}p{#1}}
\usepackage{microtype}
\usepackage{subfig}
\usepackage{tabularx}

\usepackage{enumitem,kantlipsum}

\usepackage{tikz}
\usepackage{xparse}

\NewDocumentCommand{\morpheme}{m}{%
 \tikz[baseline]{\node[draw=gray, rounded corners=2pt, inner sep=2pt, outer sep=0pt, anchor=base]{\strut #1};}%
}
\usepackage{booktabs}
\usepackage{array}
\usepackage[table]{xcolor}

\title{Scaling Subword Segmental Language Modelling}
\title{Finetuning Subword Segmentation for Low-Resource Text Generation}
\title{The Learning Dynamics of Subword Segmentation for \\ Conjunctive and Disjunctive Orthographies}
\title{The Learning Dynamics of Subword Segmentation\\ for Morphologically Diverse Languages}

\author{Francois Meyer and Jan Buys \\
  Department of Computer Science \\
  University of Cape Town \\
  \texttt{francois.meyer@uct.ac.za, jbuys@cs.uct.ac.za}}

\begin{document}
\maketitle

\begin{abstract}
Subword segmentation is typically applied in preprocessing and stays fixed during training. 
Alternatively, it can be learned \emph{during} training to optimise the training objective. In this paper we study the learning dynamics of subword segmentation: if a language model can dynamically optimise tokenisation, how do its subwords evolve during training? 
To explore this, we extend the subword segmental language model (SSLM), a framework for learning subwords during training, to support pretraining and finetuning.
We train models for three typologically diverse languages to study learning dynamics across the morphological spectrum:
Isi\-Xhosa is conjunctive (long word forms composed of many morphemes), Setswana is disjunctive (morphemes written as separate words), and English
represents a typological middle ground. 
We analyse subword dynamics from a linguistic perspective, tracking morphology, productivity, and fertility. We identify four stages of subword learning, with the morphologically complex isi\-Xhosa exhibiting greater instability. 
During finetuning, subword boundaries shift to become finer-grained.
Lastly, we show that learnable subwords offers a promising approach to improve text generation 
and cross-lingual transfer for low-resource, morphologically complex languages.

\end{abstract}

\section{Introduction}

Subword tokenisers like BPE \citep{sennrich-etal-2016-neural} and ULM \citep{kudo-2018-subword} 
are applied during preprocessing, after which they remain unchanged during language model (LM) pretraining and finetuning. If tokenisers produce subwords that are ineffective units for language modelling, this cannot be fixed during training. This is especially problematic for settings in which models are sensitive to tokenisation decisions, such as low-resource tasks \citep{zhu-etal-2019-importance, wang-etal-2021-multi-view} and morphologically rich languages \citep{zhu-etal-2019-systematic, klein-tsarfaty-2020-getting}.


\begin{figure}[t]
\vspace{-0.5cm}
  \includegraphics[trim={0 0 0 0cm},clip,width=\linewidth]{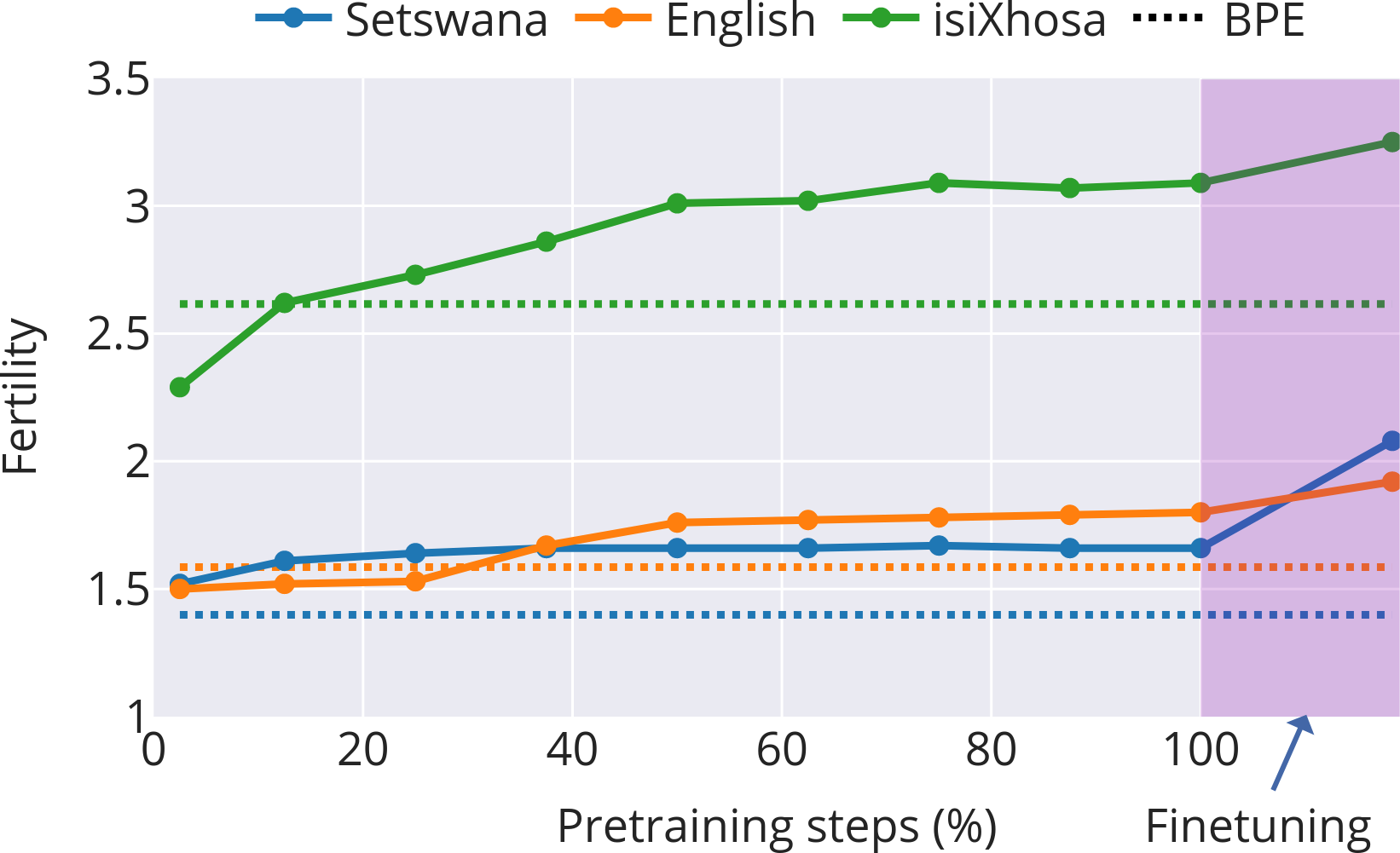}
  \caption{Subword fertility (average subwords per word) gradually plateaus for isiXhosa, while converging early for English and Setswana.} 
  \vspace{-0.375cm}
  \label{fig:fertility}
\end{figure}

Ideally, subword segmentation should be learned during end-to-end training. 
Previous works achieve this by jointly optimising segmentation and model parameters \citep{kreutzer-sokolov-2018-learning, he-etal-2020-dynamic, meyer-buys-2022-subword}. 
In this paper, we use this paradigm to study the learning dynamics of subword segmentation.
When an LM is able to learn subword segmentation, how do its subword boundaries evolve over the course of pretraining and finetuning? The question has practical implications, as it reveals how tokenisation requirements vary across different stages of training. 
There is no guarantee that subword units that are optimal for the early epochs of pretraining would also be suitable for later stages of pretraining, or meet the demands of subsequent task-specific finetuning. 

Subword learning dynamics may also depend on linguistic properties, particularly morphology. We conduct experiments on three languages spanning a range of morphological complexity: Setswana, English, and isiXhosa. Setswana and isiXhosa are South African languages that represent opposite ends of the morpho-orthographic spectrum. Both are classified as agglutinative, but they differ in how orthographic (written) words are constructed \citep{taljard-bosch-2006-comparison}. 
Isi\-Xhosa is conjunctive: morphemes are concatenated into long word forms, whose meaning depends on morphological decomposition. 
Setswana is disjunctive: morphemes are written as space-separated words and many words consist of single morphemes. English, with its limited inflectional morphology, serves as a typological middle ground. The examples in Table \ref{tab:example_phrase_segmentations} illustrate the contrasting subword structures. 



Our study requires a model capable of dynamically learning subword segmentation during pretraining and finetuning. For this purpose we extend the subword segmental language model (SSLM) \citep{meyer-buys-2022-subword}, which jointly optimises subword segmentation and LM training. 
We develop a Transformer-based variant, \textbf{T-SSLM}, incorporating learnable subword segmentation into modern LM practices like pretraining, task-specific finetuning, and text generation. 
{
\setlength{\tabcolsep}{2pt} 
\begin{table}[t]
\footnotesize
\centering
\begin{tabular}{lll}
\toprule
\textbf{Setswana} & \textbf{English}  &  \textbf{isiXhosa} \\
\midrule
\morpheme{ga} \, \morpheme{ke} \, \morpheme{itse} &
\,\,\morpheme{I} \, \morpheme{do}\morpheme{n't} \, \morpheme{know} &
\,\,\morpheme{a}\morpheme{nd}\morpheme{azi} \\
\midrule
\morpheme{ba} \, \morpheme{a} \, \morpheme{bala} &
\,\,\morpheme{they} \, \morpheme{are} \, \morpheme{read}\morpheme{ing} &
\,\,\morpheme{ba}\morpheme{funda} \\
\midrule
\morpheme{ke} \, \morpheme{a} \, \morpheme{leboga} &
\,\,\morpheme{thank} \, \morpheme{you} &
\,\,\morpheme{ndi}\morpheme{ya}\morpheme{bulela} \\
\bottomrule
\end{tabular}
\captionof{table}{Morphological decompositions of translations, illustrating a spectrum from disjunctive (Setswana) to conjunctive (isiXhosa), with English in between.}
\vspace{-0.3cm}
\label{tab:example_phrase_segmentations}
\end{table}
}

We pretrain T-SSLMs for Setswana, English, and isiXhosa, tracking learned subwords across checkpoints  
 from three linguistic perspectives. 
\textbf{Morphological alignment} measures how closely subwords align with morphological boundaries. \textbf{Productivity} \citep{gutierrez-vasques-etal-2023-languages} measures the generative capacity of subwords, 
while \textbf{Fertility} \citep{acs-2021-exploring} estimates lexical coverage. 
Across all three criteria, isiXhosa exhibits the greatest instability, while Setswana converges earliest.
We identify distinctive phases of subword learning, characterised by clear changes and trends in our linguistic metrics. 
Our results show that different stages of training and different languages necessitate characteristically different subword units.



Finally, to evaluate the impact of learnable subwords on downstream performance, we evaluate T-SSLM on isi\-Xhosa text generation. 
It comfortably outperforms tokenisation-based LMs, with isiXhosa data-to-text BLEU gains of 6.25 and a notable reduction in text degeneration. 
This suggests finetuning subword segmentation as a promising avenue for improving low-resource text generation, while confirming -- from a model performance perspective -- that subword requirements evolve across different stages of training.





\section{Related Work}


\subsection{Learning dynamics of language models} \label{relatedwork_taskadaptive}

Previous research tracked the evolution of syntactic and semantic knowledge during training. Some studies track internal representation change \citep{saphra-lopez-2019-understanding, chiang-etal-2020-pretrained, liu-etal-2021-probing-across, muller-eberstein-etal-2023-subspace}, while others track 
performance on grammatical tasks \citep{choshen-etal-2022-grammar, evanson-etal-2023-language}. In all these studies (as in most LM research) tokenisation is fixed, so there is no subword learning trajectory to study. 
\citet{gutierrez-vasques-etal-2023-languages} 
analyse how BPE token properties change over successive merges; however, BPE is still fixed after preprocessing, so the emergence of subwords in LM training is not studied. 
That requires considering approaches that incorporate subword learning into training.

\subsection{Unifying training and tokenisation}

Previous works unify training and subword learning by marginalising over different candidate tokenisations during training. This has been used to learn subwords for translation \citep{kreutzer-sokolov-2018-learning, he-etal-2020-dynamic, meyer-buys-2023-subword} and unsupervised Chinese word segmentation \citep{sun-deng-2018-unsupervised, kawakami-etal-2019-learning, downey-etal-2022-masked}. 
\citet{meyer-buys-2022-subword} adapt this technique to propose subword segmental language modelling (SSLM), an LSTM-based LM architecture that dynamically optimises subword segmentation. They train small-scale SSLMs for low-resource, agglutinative languages. 
They do not conduct finetuning or downstream evaluation, but perplexity-based evaluation demonstrates the potential of SSLM over tokenisation-based LMs.

\section{Transformer SSLM}
\label{sec:transformer_sslm}


We leverage the technique underlying the SSLM of \citet{meyer-buys-2022-subword} -- learning subword units during LM training by marginalising over potential tokenisations -- as an opportunity to study subword learning dynamics. The original SSLM architecture is based on a shallow LSTM. 
To align SSLM with current best practices for pretraining, task-finetuning, and open-ended text generation, we develop a Transformer version of the model. We propose \textbf{T-SSLM}, a Transformer-based adaptation of SSLM that brings the model up to date with modern LM architectures. 

From a modelling perspective, our T-SSLM is a conceptually straightforward but technically non-trivial extension of the original SSLM. We do not position our architectural changes as the main contribution of this paper. Our primary research goal is empirical analysis: to study subword learning dynamics across different stages of training. 
We develop T-SSLM as a new tool for studying subword segmentation as a learnable component of language modelling. Our implementation of T-SSLM is publicly available.\footnote{\url{https://github.com/francois-meyer/transformer-sslm}}

In this section, we present a technical overview of how T-SSLM dynamically learns subword segmentation during pretraining and finetuning. 
For a comprehensive introduction to the underlying SSLM framework, we refer readers to \citet{meyer-buys-2022-subword}.
Here we summarise the method behind the original SSLM, outline how we adapt it to Transformer-based modelling, and explain how we use T-SSLM to track subword learning. 

\subsection{SSLM background}

For a standard tokenisation-based LM with parameters $\theta$, the probability assigned to a document $D$ is based on a particular fixed subword tokenisation $T$ (e.g. obtained with BPE), computed using the chain rule as
\begin{align} \label{tokenised_formula} 
p_{\theta}(D, T) = \prod_{i=1}^{n} p_{\theta} (t_i | t_{<i}),
\end{align}
where $t_1, t_2, ..., t_n$ is the sequence of subwords that $D$ is tokenised into by $T$.

Instead of using a single pre-determined tokenisation $T$, SSLM marginalises over all possible subword segmentations during training. The probability assigned to a document $D$ is computed as
\begin{align} \label{marginalised_formula} 
p_{\theta}(D) = \sum_{T \in \pi(D)}p_{\theta}(D, T), 
\end{align}
where $\pi(D)$ denotes the set of all potential subword segmentations of document $D$. 

The number of possible subword segmentations of a document $D$ grows exponentially with its length, so the marginalisation in Eq.~\ref{marginalised_formula} becomes intractable. \citet{meyer-buys-2022-subword} propose a dynamic programming algorithm to compute Eq.~\ref{marginalised_formula} efficiently. 
Suppose document $D$ consists of characters $\mathbf{c} = c_1, c_2, \hdots, c_{|D|}$. The algorithm computes the marginalised probability up to the $k^\mathrm{th}$ character as $p(D_{1:k}) = \alpha_k$, where $\alpha_0 = 1$ and $\alpha_k$ are forward scores computed as 
\begin{align} \label{dp}
    \alpha_k = \sum_{j= f(c, k)}^{k} \alpha_{j-1} p(t = c_{j:k} | c_{<j}), 
\end{align}
where $f$ returns the index of the first character in the word containing the $k^{\mathrm{th}}$ character, thereby prohibiting subwords that cross word boundaries. During pretraining, the marginal (Eq. \ref{marginalised_formula}) of the full document is optimised by computing $p(D) = \alpha_{|D|}$.

\subsection{T-SSLM: adapting SSLM for pretraining and finetuning}

\citet{meyer-buys-2022-subword} parameterise the SSLM described above with an LSTM \citep{hochreiter-schmidhuber-1997-long}, so the subword probabilities in Eq.~\ref{dp} are conditioned on LSTM encodings. 
We propose T-SSLM, a new variant of SSLM parameterised by a Transformer \citep{vaswani-etal-2017-attention}. This enables larger-scale pretraining and aligns SSLM with modern LM architectures.

To construct a neural LM that can compute $p_{\theta}(D, T)$ for any candidate segmentation $T$, \citet{meyer-buys-2022-subword} introduce several architectural innovations, which we replicate in T-SSLM. One of these innovations is to encode the preceding text $t_{<i}$ as an untokenised character sequence $c_{<j}$ to achieve computationally feasible conditioning on the tokenisation history. In T-SSLM, we use a character-level Transformer for this encoding instead of an LSTM. This Transformer forms the backbone of our architecture: the subword probabilities computed in Eq.~\ref{dp} are conditioned on its learned output representations. 

Like the original SSLM, T-SSLM computes the next-subword probability $p_{\theta} (t_i | \cdot)$ as a weighted mixture of two distributions: one over character sequences (\( p_{\mathrm{char}} \)) and another over a subword lexicon (\( p_{\mathrm{lex}} \)). The lexicon contains a fixed set of high-frequency subwords, while \( p_{\mathrm{char}} \) allows the model to compose arbitrary subwords from characters. This design, with implementation details in Appendix~\ref{appendix_mixture_subsection}, enables T-SSLM to adapt both its segmentation algorithm \emph{and} its subword vocabulary over the course of training.

The Transformer backbone enables scalable pretraining. In addition, we introduce two extensions to the original SSLM to enable downstream application of T-SSLM. First, we modify the dynamic programming algorithm. 
The original formulation of Eq.~\ref{dp} computes the marginal likelihood \( p(D) \) of a full sequence \( D \), but prompt-based finetuning requires maximising the \emph{conditional} likelihood \( p(O | C) \), where \( C \) is an input prompt and \( O \) is the expected completion. 
We derive a modified version of the original dynamic program to handle conditional likelihoods (see Appendix~\ref{appendix_finetuning_subsection} for derivation).
Second, we develop a custom decoding algorithm (outlined in Appendix~\ref{appendix_decoding_subsection}) to enable open-ended text generation with T-SSLM, since standard beam search implementations are incompatible with the mixture model described above.  


\begin{figure*}[t]
    \subfloat[Setswana]{\includegraphics[width=5.2cm]{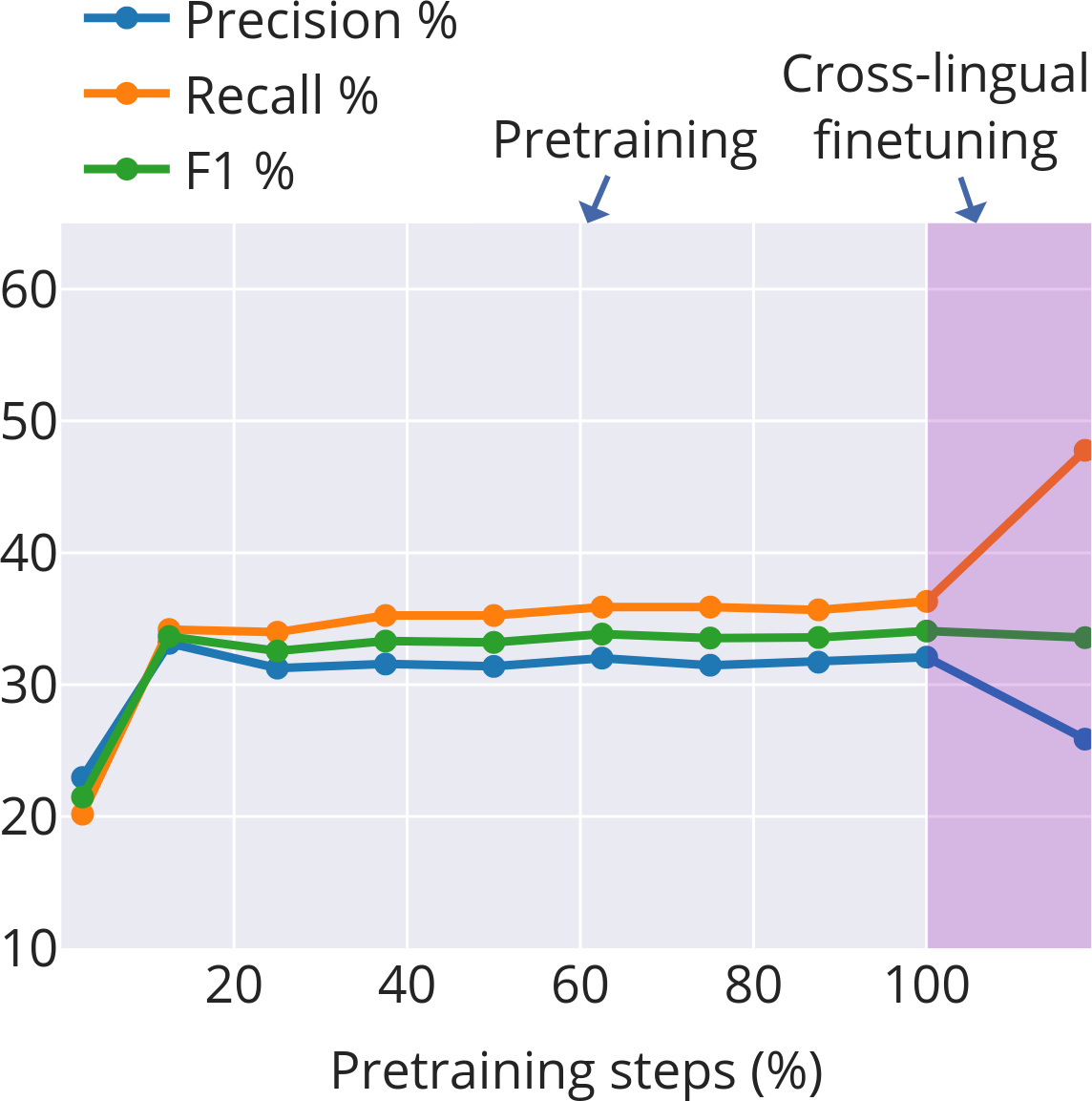}\label{fig:ums:ts}}%
     \,\,
    \subfloat[English]{\includegraphics[width=5.2cm]{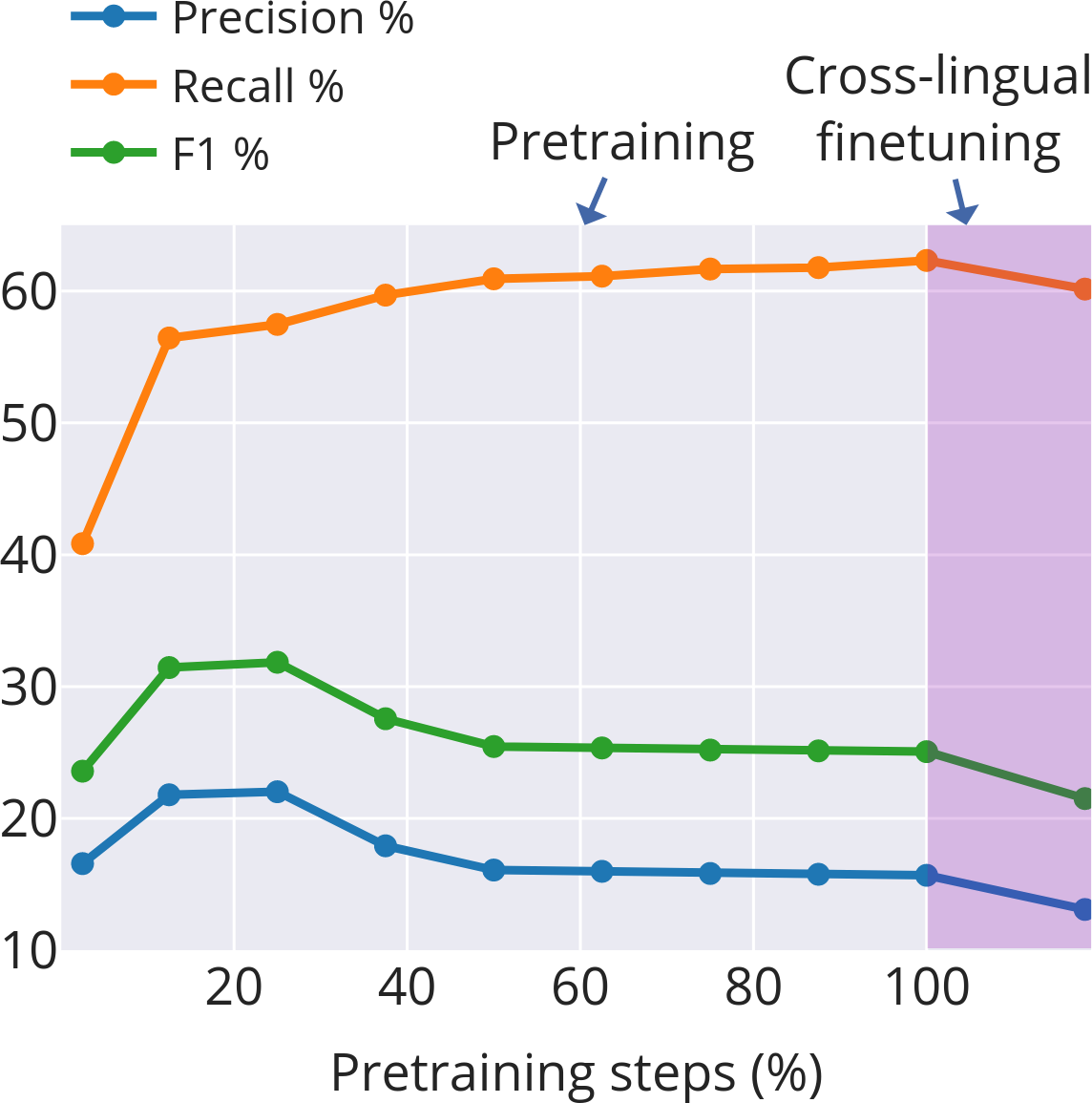}\label{fig:ums:en}}%
    \,\,
    \subfloat[isiXhosa ]{\includegraphics[width=5.2cm]{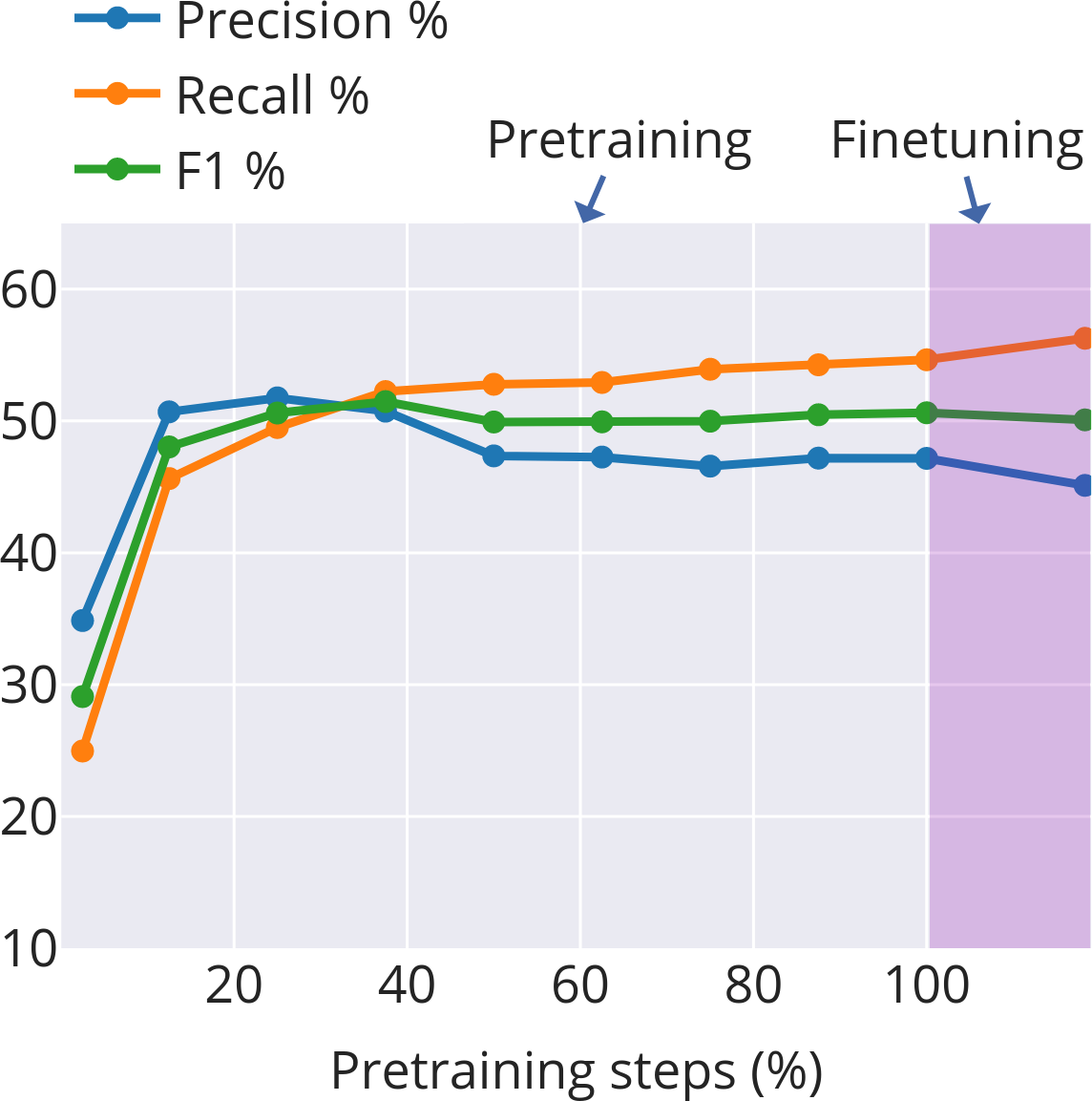}\label{fig:ums:xh}}%
    \vspace{-0.1cm}
    \caption{Morphological boundary overlap between learned subwords and morphological segmentations. Pretraining is performed on Setswana/English/isiXhosa, while finetuning is performed on isiXhosa data-to-text.}%
\vspace{-0.2cm}
    \label{fig:morph_plots}%
\end{figure*}

\subsection{Extracting learned subwords for analysis}


During pretraining, T-SSLM learns to assign higher probability to segmentations $T$ that maximise Eq.~\ref{marginalised_formula}. After pretraining, or for any intermediate checkpoint, we can extract the model's preferred segmentation $T^*$ of document $D$
by applying Viterbi decoding to the dynamic program of Eq.~\ref{dp}:
\begin{align}
T^* = \argmax_{T \in \pi(D)} p_\theta(D, T). \label{eq:viterbi}
\end{align} 
This is the probability-maximising segmentation according to the model's current parameters $\theta$, reflecting the subword segmentation that the model has learned as optimal at that stage of training.

During finetuning, T-SSLM adapts its subword segmentation to suit the requirements of the downstream task. Rather than optimising full-document likelihood, as in pretraining, the model now maximises the conditional likelihood of output completions given prompts. 
This leads to new subword preferences that better support task-specific generation. We again use the Viterbi algorithm to extract the model's preferred segmentation of a document after finetuning, revealing how subword segmentation changes from pretraining to finetuning.

\section{Experimental Setup}

We pretrain monolingual T-SSLMs for Setswana, English, and isiXhosa. 
Pretraining and finetuning hyperparameters are provided in Appendix \ref{appendix_hyperparameters}. 

\subsection{Pretraining}
T-SSLM's time complexity (detailed in \S \ref{computational_complexity_subsection}) makes it computationally challenging to scale the pretraining data size to match state-of-the-art LMs.
However, we view the technique as an opportunity for analysis: a unique lens through which to analyse the role of subword segmentation in LM training. 
For pretraining, we prioritise high-quality corpora over scale to ensure that models are trained on clean, representative language input. Our corpus sizes are listed in Table \ref{table_datasetstats}. 
For isi\-Xhosa we use the 13m-word WURA \cite{oladipo-etal-2023-better} corpus, created by filtering mC4 \cite{xue-etal-2021-mt5} to discard low-quality data. For the lower-resourced Setswana we use the 4m-word PuoData corpus \citep{marivate2023puoberta}. For English we use the 10m-word BabyLM corpus \citep{warstadt-etal-2023-findings}. 
To track learning dynamics, we store model checkpoints at regular pretraining intervals. After pretraining, we analyse the subwords of nine checkpoints spread evenly across pretraining: one checkpoint after the first epoch, one after the final epoch, and seven additional checkpoints spaced at regular intervals (1/8ths of total pretraining).

\begin{table}[t] \small
    \centering
    \begin{tabular}{lccc}
    \toprule
        \textbf{Dataset} & \textbf{Train} & \textbf{Valid} & \textbf{Test} \\  
         \midrule
         \multicolumn{4}{c}{\textbf{Pretraining (\# words)}} \\
         \midrule
          Setswana: PuoData & 4.06m & 247k & 212k \\
          English: BabyLM & 9.95m & 998k & 994k \\
    isiXhosa: WURA & 13.08m & 767k & 724k \\ 
    \midrule
         \multicolumn{4}{c}{\textbf{Downstream isiXhosa finetuning (\# examples)}} \\
         \midrule
    Triples-to-isiXhosa (T2X) & 3,859 & 600 &   378 \\
    MasakhaNEWS (MNH)  & 1,032& 147& 297  \\
    \bottomrule
    \end{tabular}
    \caption{Pretraining and finetuning dataset statistics.}
    \label{table_datasetstats}
    \vspace{-0.2cm}
\end{table}

\subsection{Finetuning}
\label{sec:finetuning}

We study how subwords change during finetuning from two perspectives: how they adapt to a specific downstream task, and how they shift when transferring across languages with different morphological structures (i.e. cross-lingual finetuning). We finetune all our T-SSLMs (Setswana, English, isiXhosa) on two isiXhosa text generation tasks, data-to-text and headline generation. 
We focus on isiXhosa for finetuning because its complex morphological structure makes it a challenging language to adapt subword modelling cross-lingually. 

For data-to-text we use the T2X dataset \citep{meyer-buys-2024-triples}. It contains triples (subject, relation, object) paired with descriptive isi\-Xhosa sentences e.g. (South Africa, leader, Cyril Ramaphosa) $\rightarrow$ ``uCyril Ramaphosa yinkokheli yoMzantsi Afrika'' (``Cyril Ramaphosa is the leader of South Africa''). For headline generation from the body of a news article we use MasakhaNEWS \citep{adelani-etal-2023-masakhanews}, which contains news articles paired with headlines. 
This is a much harder task than data-to-text, especially considering the smaller finetuning dataset (see Table \ref{table_datasetstats}). 

\subsection{Baselines} \label{subsection_baselines}

For each language, we pretrain and finetune five LMs with fixed subword tokenisation (hyperparameters are reported in Appendix \ref{appendix_baseline_hyperparams}). We use the same datasets as our T-SSLMs to ensure a fair comparison.  
Three baselines use well-established tokenisers: (1) BPE \citep{sennrich-etal-2016-neural}, (2) ULM \citep{kudo-2018-subword}, and (3) BPE-dropout \citep{provilkov-etal-2020-bpe}.
Additionally, we use (4) character-level segmentation and (5) byte-based modelling, both of which have been shown to outperform subword-based LMs on low-resource languages \citep{edman-etal-2022-subword, edman-etal-2024-character, adelani-etal-2022-thousand}. 

These LMs are baselines for downstream evaluation (Section \ref{sec:nlg}), allowing us to compare learnable subword segmentation and fixed tokenisation in terms of text generation performance. Besides plotting BPE as a reference in Figure \ref{fig:fertility}, we do not include these baselines in our analysis of subword learning dynamics (Section \ref{sec:subword_learning_dynamics}), since their tokenisation is fixed and does not evolve during training. 

\subsection{Subword analysis}  \label{subsection_analysis}

For any given T-SSLM checkpoint, we can encode a document and extract its probability-maximising subword segmentation using the Viterbi algorithm. This reveals the model's preferred tokenisation of the text at that point in pretraining/finetuning. We analyse and track these learned segmentations from three linguistic perspectives.

\subsubsection{Morphological alignment} \label{subsection:ums}

Our models do not have access to morphological annotations, but we investigate to what extent they ``discover'' morphemes as subword units by tracking the amount of overlap between learned subwords and morphemes. For Setswana and isiXhosa we use the NCHLT test corpus \citep{eiselen-puttkammer-2014-developing}, which contains 5,000 morphologically annotated words for each language. For English we use the 2022 SIGMORPHON dataset \citep{batsuren-etal-2022-sigmorphon}, which contains 18,000 morphologically segmented words. We compare the subwords learned by our models to the gold-standard morphological segmentations. To measure morphological alignment, we compute precision, recall, and F1 for morphological boundary identification.


\begin{figure*}[t]%


    \subfloat[Setswana]{\includegraphics[width=5.2cm]{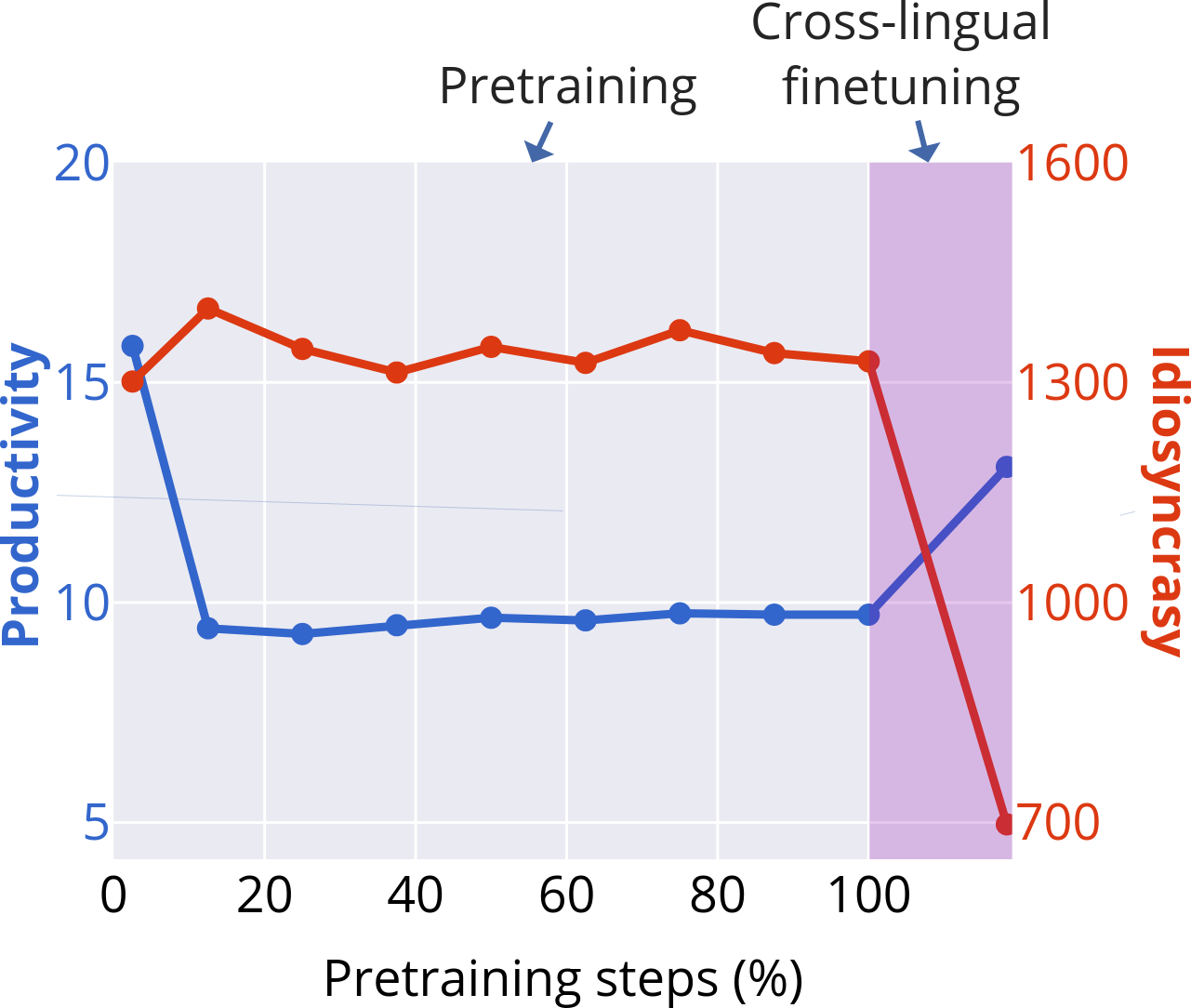}\label{fig:productivity:ts}}%
    \,\,
    \subfloat[English]{\includegraphics[width=5.2cm]{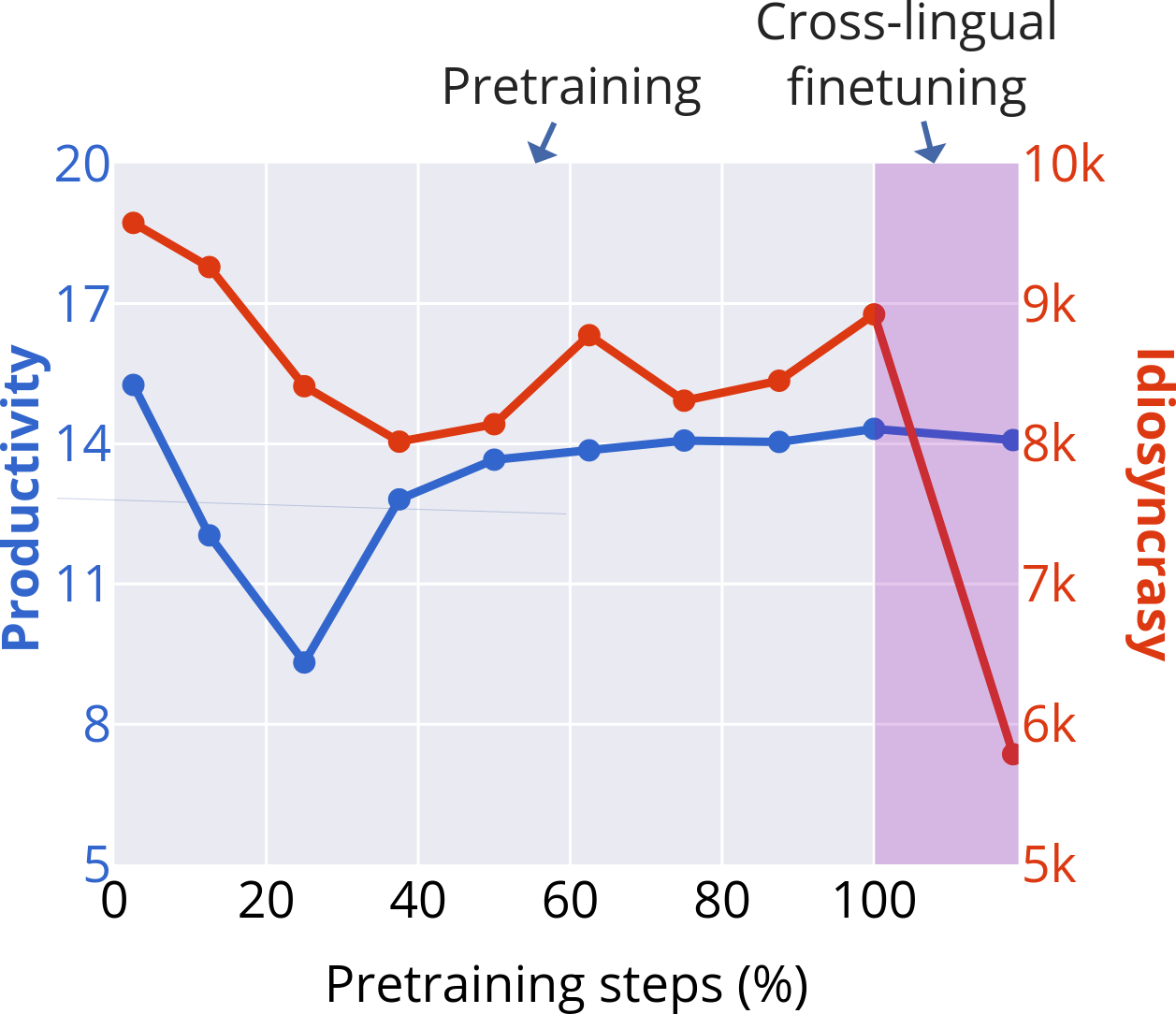}\label{fig:productivity:en}}%
    \,\,
    \subfloat[isiXhosa]{\includegraphics[width=5.2cm]{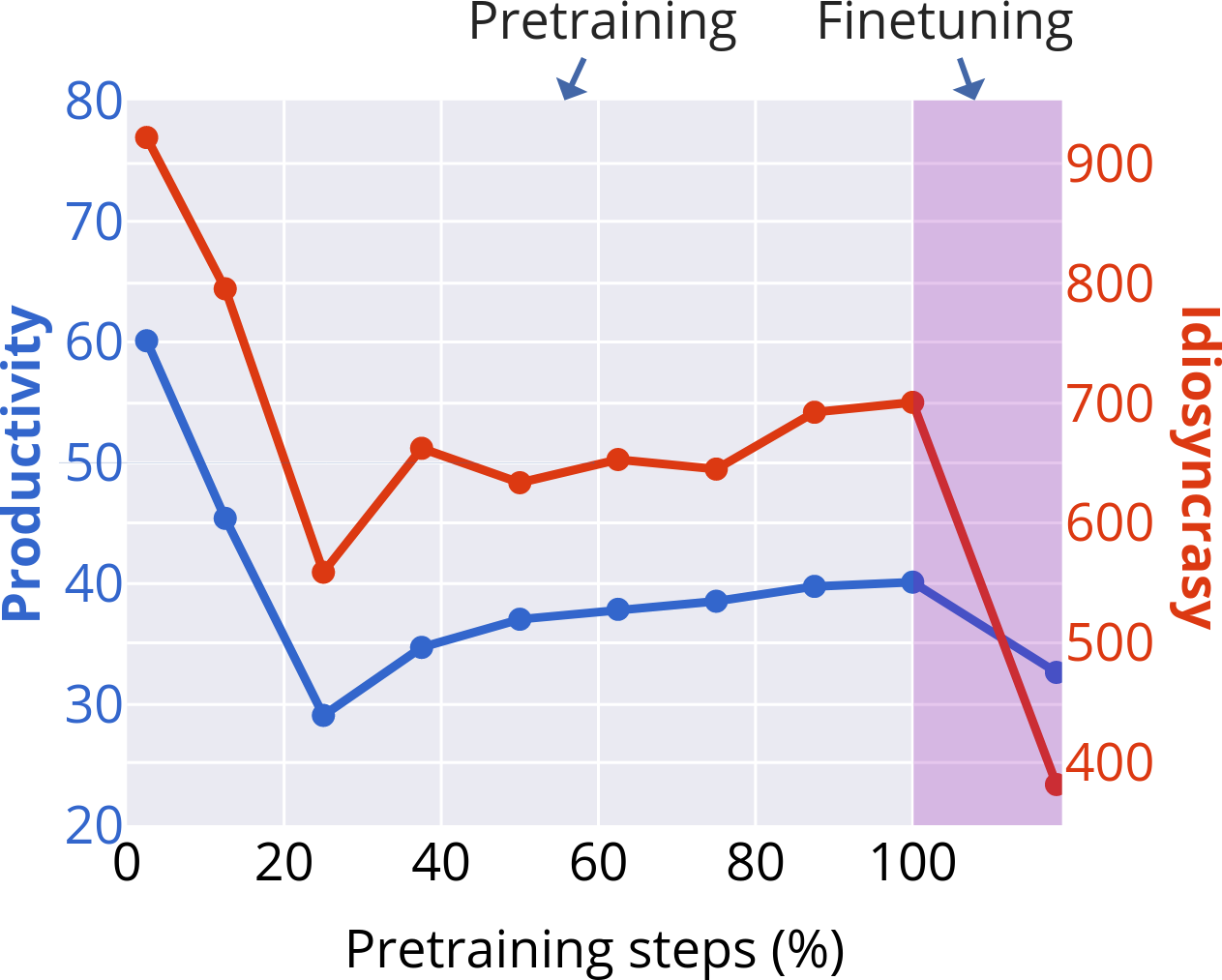}\label{fig:productivity:xh}}%

    \vspace{-0.1cm}
    \caption{The average productivity and idiosyncrasy of learned subwords. Pretraining is performed on isiXhosa and Setswana, respectively, while finetuning is conducted on isiXhosa data-to-text generation.}%
    \vspace{-0.2cm}
    \label{fig:productivity}%
\end{figure*}

\subsubsection{Productivity and Idiosyncrasy}

Morphological productivity is the capacity of a morpheme to combine with other morphemes to form new words \citep{Bybee_2001}. 
Productive morphemes occur in many word types, often forming regular morphological patterns. For example, in isi\-Xhosa prefixes can be attached to stems to indicate plurality 
e.g. ``\underline{aba}hlobo'' (friend\underline{s}) and ``\underline{aba}zali'' (parent\underline{s}). 
In contrast, \emph{idiosyncratic} morphemes are contained in few word types, but these types occur frequently. For example, the isi\-Xhosa stem ``--ntu'' (human) will not occur in as many unique words as prefixal morphemes, but the words in which ``--ntu'' occurs are commonly used e.g. ``um\underline{ntu}'' (person) and ``aba\underline{ntu}'' (people). 

We quantify productivity and idiosyncrasy as proposed by \citet{gutierrez-vasques-etal-2023-languages}, 
\begin{align} \label{productivity} 
\text{productivity}(s) &= |W_s|, \\
\text{idiosyncrasy}(s) &= \frac{\sum_{w \in W_s} \text{freq}(w)}{|W_s|},
\end{align}
where $W_s$ is the set of unique words in which $s$ occurs and $\text{freq}(w)$ is the corpus frequency of a word $w$. Productivity quantifies how many \emph{types} contain a subword, while idiosyncrasy is the average \emph{token} frequency of types containing a subword. 
We compute these metrics from the subword segmentation and frequencies of our LM validation corpora.


\subsubsection{Fertility} The fertility of a tokeniser is the average number of subwords per segmented word \citep{acs-2021-exploring}. It is a measure of lexical coverage -- the extent to which the words of a language are included in the tokeniser vocabulary. A fertility of 1 indicates that every word in a corpus is included in the vocabulary (no segmented words), while a higher fertility corresponds to finer-grained segmentation and lower lexical coverage. We also compute fertility based on the segmentation of our LM validation corpora.

\section{Subword Learning Dynamics} \label{sec:subword_learning_dynamics}

We analyse how T-SSLM subword units evolve during pretraining and finetuning.\footnote{We present the analysis for finetuning on isi\-Xhosa data-to-text. Results are similar for our other task, headline generation.} 
The learning dynamics are visualised in Figures \ref{fig:fertility}, \ref{fig:morph_plots} and \ref{fig:productivity}. 
We identify four distinctive stages of subword learning. We include representative examples of segmentations at each stage in Tables \ref{table_examples} and \ref{table_fer_examples}.

\subsection*{Stage 1: Rapid evolution of subwords}

%
The initial stage of subword learning (first 20\% of pretraining) is characterised by dramatic changes in subword boundaries. 
The models start with randomly initialised subword boundaries and find subword units that optimise the training objective in the initial period of a high-loss training. 
As shown in Figure \ref{fig:morph_plots}, initial subwords have little morphological alignment, but models quickly adjust their subword boundaries to more closely align with morpheme boundaries.
This shows that an important part of the early subword learning is the emergence of morphemes, to an extent, as subword units. 
We include examples of morphological subwords emerging after stage 1 in Table \ref{table_examples}.

\newcommand{\morphemegreen}[1]{\tikz[baseline]{\node[draw=green!50!black, fill=green!10, rounded corners=2pt, inner sep=2pt, outer sep=0pt, anchor=base]{\strut #1};}}
\newcommand{\morphemeplain}[1]{%
  \tikz[baseline]{\node[draw=gray, rounded corners=2pt, inner sep=2pt, outer sep=0pt, anchor=base]{\strut #1};}%
}
{
\setlength{\tabcolsep}{4.75pt} 
\begin{table*}[t] \small
    \centering
    \renewcommand{\arraystretch}{1.3}
    \begin{tabular}{l|ccc|ccc|ccc}
    \toprule
        \textbf{Language} & \multicolumn{3}{c|}{\textbf{Setswana}}  & \multicolumn{3}{c|}{\textbf{English}}  & \multicolumn{3}{c}{\textbf{isiXhosa}} \\
\midrule
\textbf{Word} & malapa & dingwe & rona & assuming & lately & donuts & umntu & ukuxhasa & zoncedo\\  
\midrule


\textbf{Stage 1} 
& \morphemeplain{mal}\morphemeplain{apa} &  \morphemeplain{dingw}\morphemeplain{e} & \morphemegreen{rona} & \morphemeplain{assu}\morphemeplain{ming} & \morphemeplain{l}\morphemeplain{ately} & \morphemeplain{don}\morphemeplain{uts} & \morphemeplain{umntu} & \morphemeplain{uku}\morphemeplain{xhasa} & \morphemeplain{zon}\morphemeplain{cedo}\\

\textbf{Stage 2} 
& \morphemegreen{ma}\morphemeplain{lapa} &  \morphemegreen{di}\morphemeplain{ngwe} & \morphemegreen{rona} & \morphemeplain{a}\morphemeplain{ssum}\morphemegreen{ing} & \morphemegreen{late}\morphemegreen{ly} & \morphemeplain{donut}\morphemegreen{s} & \morphemeplain{um}\morphemegreen{ntu} & \morphemeplain{uku}\morphemegreen{xhas}\morphemegreen{a} & \morphemeplain{zoncedo}\\

\textbf{Stage 3} 
& \morphemegreen{ma}\morphemeplain{lapa} & \morphemegreen{di}\morphemeplain{ngwe} & \morphemegreen{rona} & \morphemeplain{a}\morphemeplain{ssum}\morphemegreen{ing} & \morphemeplain{latel}\morphemeplain{y} & \morphemeplain{donut}\morphemegreen{s} & \morphemegreen{u}\morphemegreen{m}\morphemegreen{ntu} & \morphemeplain{uku}\morphemegreen{xhas}\morphemegreen{a} & \morphemegreen{zo}\morphemeplain{ncedo}\\

\textbf{Stage 4} 
& \morphemegreen{ma}\morphemeplain{lapa} &  \morphemeplain{dingw}\morphemeplain{e} & \morphemegreen{rona} & \morphemeplain{assu}\morphemeplain{m}\morphemegreen{ing} & \morphemeplain{latel}\morphemeplain{y}  & \morphemeplain{donut}\morphemegreen{s} & \morphemeplain{um}\morphemegreen{ntu}  & \morphemeplain{uku}\morphemeplain{xhasa} & \morphemegreen{zo}\morphemeplain{ncedo}\\

    \bottomrule
    \end{tabular}
    \vspace{-0.1cm}
    \caption{Examples of changes in learned subword segmentations throughout training, highlighting subwords corresponding to morphemes (morphemes obtained from linguistically annotated datasets introduced in \S\ref{subsection:ums}).}
    \label{table_examples}
    
\end{table*}
}


\begin{table*}[t] \small
    \centering
    \begin{tabular}{l|l|l|l|l} 
    \toprule
         
\textbf{Pretrained} 
& \morpheme{A}\morpheme{keem} \, \morpheme{D}\morpheme{ent} 
& \morpheme{zi}\morpheme{se}\morpheme{New} \, \morpheme{Y}\morpheme{ork} 
& \morpheme{M}\morpheme{e}\morpheme{x}\morpheme{ico} &  \morpheme{sase}\morpheme{L}\morpheme{o}\morpheme{s} \, \morpheme{A}\morpheme{ngele}\morpheme{s}
 \\
\midrule

\textbf{Finetuned}  
& \morpheme{A}\morpheme{k}\morpheme{e}\morpheme{e}\morpheme{m} \, \morpheme{D}\morpheme{e}\morpheme{n}\morpheme{t} 
& \morpheme{zi}\morpheme{se}\morpheme{N}\morpheme{e}\morpheme{w} \, \morpheme{Y}\morpheme{o}\morpheme{r}\morpheme{k} 
& \morpheme{M}\morpheme{e}\morpheme{x}\morpheme{i}\morpheme{c}\morpheme{o}
& \morpheme{sase}\morpheme{L}\morpheme{o}\morpheme{s} \, \morpheme{A}\morpheme{n}\morpheme{g}\morpheme{e}\morpheme{l}\morpheme{e}\morpheme{s}
\\





        \bottomrule
    \end{tabular}
    \vspace{-0.1cm}
     \captionof{table}{Segmentations of entities in the data-to-text dataset, produced by isiXhosa T-SSLM before/after finetuning. Fertility increases as named entities, which lack meaningful subword units, are segmented into characters.}   
    \label{table_fer_examples}
    \vspace{-0.3cm}
\end{table*}

Figure \ref{fig:productivity} shows similarly rapid changes in subword productivity. For English and isiXhosa, productivity and idiosyncrasy decrease at a constant rate. \citet{gutierrez-vasques-etal-2023-languages} found a similar pattern in early BPE merges. Highly productive, high-frequency subwords are identified at the outset, after which the model incorporates progressively less common subwords. This leads to decreased productivity and idiosyncrasy. 
For Setswana, the learning trajectory is different. Productivity also decreases, but idiosyncrasy remains stable. This also agrees with \citet{gutierrez-vasques-etal-2023-languages}, who showed that BPE tokens for morphological simple languages have low productivity but high idiosyncrasy.  
Many Setswana subwords are near-whole words, so they occur in few types, but the types are not rare (i.e. high idiosyncrasy). 

\subsection*{Stage 2: Inflection point for English \& isiXhosa}

After an initial period of adjustment, Setswana subwords stablise.
In contrast, at this stage (between 20\% and 40\% of pretraining) isiXhosa and English undergo a sudden change in learning trajectories. After the initial period of decreasing productivity and idiosyncrasy, both metrics increase (Figure \ref{fig:productivity:xh}). This also coincides with a large increase in fertility (Figure \ref{fig:fertility}), indicating finer grained segmentation. We interpret this as the initially expanded vocabulary being trimmed down. In the initial phase of exploration, the model identified a broad set of candidates for effective subwords. Subsequently, learning shifts to finding subwords that improve LM generalisation, by preferring subwords that occur in many word types (e.g. productive morphemes) and/or in high-frequency words (idiosyncratic). 

This inflection point represents a shift in the morphological alignment of isiXhosa subwords, as shown by Figure \ref{fig:ums:xh}. Initially a large proportion of subwords have morphological boundaries (high precision). After the inflection point, the model increases its coverage of morphological boundaries (high recall), but at the cost of more morphologically unsound boundaries. The model is erring on the side of overly-aggressive segmentation, as shown in the finer-grained segmentation of ``umntu'' and ``zoncedo'' in Table \ref{table_examples} (from stage 2 to 3). 
This increases morphological boundary coverage but over-segments some morphemes. 

\subsection*{Stage 3: Stabilised learning dynamics}

The mid-to-late phases of pretraining are more stable than early subword learning. 
Throughout pretraining, isiXhosa and English dynamics are more variable than Setswana. For example, Setswana fertility quickly converges to a low value of 1.66.  
Because of its disjunctive orthography, Setswana 
leaves little opportunity for aggressive subword segmentation, so low fertility is unsurprising. 
English, which exhibits some morphological inflection, settles on a slightly higher fertility of 1.8. 
In contrast, isiXhosa fertility gradually increases to 3.05, almost double that of Setswana. 

The unstable dynamics of isi\-Xhosa demonstrates the challenges of modelling the subword structure of a morphologically complex language. Even late in pretraining, optimising the loss requires changes in subword boundaries, as better subword units continually emerge. In contrast, the Setswana subwords discovered early on are sufficient for the remainder of pretraining.

Figure \ref{fig:fertility} plots fertility of BPE tokenisers for each language (from the baselines of \S\ref{subsection_baselines}) as fixed references. For each language, the fertility of learned subwords is slightly higher than BPE. Equipped with the ability to optimise subword boundaries, T-SSLM learns finer-grained segmentation than BPE. One possibility is that T-SSLM converges on tokenisation that balances the benefits of BPE and character-level modelling, suggesting a potentially more optimal middle ground between them.

\newlength{\colwidth}
\setlength{\colwidth}{0.385cm}

\newcolumntype{L}{>{\centering\arraybackslash}p{0.4cm}} 
\newcolumntype{C}{>{\centering\arraybackslash}p{0.57cm}} 
\newcolumntype{R}{>{\centering\arraybackslash}p{0.5cm}} 

\begin{table*}[t] 
	\footnotesize
	\centering
	\begin{tabular}{l|LCR|LCR|LCR|LCR|LCR}
		\toprule
          \textbf{Pretrained $\rightarrow$} & \multicolumn{3}{c|}{\textbf{isiXhosa}} & \multicolumn{3}{c|}{\textbf{English}} & \multicolumn{3}{c|}{\textbf{Setswana}} &\multicolumn{3}{c|}{\textbf{isiXhosa}} & \multicolumn{3}{c}{\textbf{English}}  \\

\midrule
           \textbf{Finetuned $\rightarrow$}  & \multicolumn{9}{c|}{\textbf{isiXhosa data-to-text }}& \multicolumn{6}{c}{\textbf{isiXhosa headline generation}}\\
\midrule
	\textbf{Model} & chrF & \scalebox{.85}[1.0]{BLEU} &deg\% & chrF & \scalebox{.85}[1.0]{BLEU} &deg\%& chrF & \scalebox{.85}[1.0]{BLEU}&deg\%& chrF & \scalebox{.85}[1.0]{BLEU} &deg\%& chrF & \scalebox{.85}[1.0]{BLEU} &deg\%\\
       \midrule
BPE&	44.9&	12.0&	6.4&	41.1&	10.1&	\textbf{1.6}&	41.8&	10.3&	3.7&	14.9&	0.0&	10.8&	12.3&	0.2&	1.0\\
ULM&	46.1&	13.2&	6.1&	36.5&	8.2&	2.1&	42.5&	11.5&	\textbf{1.1}&	15.6&	0.0&	8.1&	12.3&	0.2&	2.0\\
BPE-dropout&	44.6&	11.0&	4.5&	35.6&	7.4&	2.7&	39.9&	9.8&	1.9&	14.4&	0.0&	8.4&	13.9&	0.2&	2.0\\
Characters&	43.5&	7.6&	5.3&	49.0&	10.0&	2.4&	44.5&	7.3&	4.8&	17.1&	0.8&	2.4&	17.3&	0.2&	\textbf{0.7}\\
Bytes&	48.2&	13.2&	4.5&	48.7&	14.3&	3.7&	45.1&	10.1&	7.9&	18.0&	0.0&	5.1&	18.5&	0.2&	1.0 \\
T-SSLM&	\textbf{49.2}&	\textbf{19.5}&	\textbf{1.1}&	
        \textbf{49.1}&	\textbf{17.0}&	\textbf{1.6}&	
        \textbf{46.3}&	\textbf{14.4}&	2.4&	
        \textbf{21.7}&     \textbf{1.9}&	\textbf{0.3}&	
        \textbf{21.6}&	\textbf{0.9}&	3.4\\

		\midrule

	\end{tabular}
    \vspace{-0.1cm}
	\caption{Test set performance of finetuned LMs on isiXhosa tasks, measured by BLEU and chrF (normalised between 0 and 100) and text degeneration (deg\% is the proportion of examples that produce incoherent, repetitive text). We omit headline generation results for Setswana-pretrained models, as all models produced zero scores.} 	\label{results_isixhosa_nlg}
    \vspace{-0.3cm}
\end{table*}

\subsection*{Stage 4: Task-oriented realignment}

During finetuning, fertility increases for all three languages (Figure \ref{fig:fertility}).
Previous work showed finer-grained units, such as characters or bytes, are beneficial for low-resource languages with small finetuning datasets \citep{edman-etal-2022-subword, edman-etal-2024-character, adelani-etal-2022-thousand}. 
More aggressive segmentation may also suit the data-to-text dataset, since it contains many named entities (e.g. ``Akeen Dent'' in Table \ref{table_fer_examples}). Names cannot be segmented into semantically coherent subwords, since they do not consist of morphemes. It is reasonable to model names as character sequences, which is what we observe after finetuning. We present examples of this change from pretraining to finetuning in Table \ref{table_fer_examples}. 

For isi\-Xhosa, productivity and idiosyncrasy decrease during finetuning (Figure \ref{fig:productivity:xh}). 
As the model narrows in on subword patterns that occur frequently in the data-to-text dataset, its subword units lose the more general morphological expressivity required for isiXhosa pretraining. 
Our Setswana and English T-SSLMs are finetuned cross-lingually on isiXhosa data-to-text. To model the complex morphological structure of isiXhosa, they lose some morphological alignment in the language of pretraining (Figure \ref{fig:ums:ts}) and improve isi\-Xhosa morphological alignment (Figure \ref{fig:xl-ums} in the appendix). 


\section{Downstream Text Generation} \label{sec:nlg}

Next we study subword learning dynamics via its effect on downstream performance.
Our preceding analysis found that finetuning subword segmentation leads to the emergence of task-specific subwords (examples in Table \ref{table_fer_examples}). This raises the question of whether the ability to adapt subword boundaries during finetuning provides a performance advantage over models with fixed tokenisation (e.g. BPE or ULM), which lack this flexibility.

We finetune T-SSLM and five tokenisation-based baseline LMs (see \S\ref{subsection_baselines}) on two isi\-Xhosa text generation tasks (data-to-text and headline generation, see \S\ref{sec:finetuning}).
We evaluate performance with BLEU \citep{papineni-etal-2002-bleu} and chrF \citep{popovic-2015-chrf}, 
which quantifies subword-level overlap.
Additionally we measure the proportion of test examples that produce degenerative text (incoherent or repetitive language). A generation is considered degenerative if a word is repeated at least three times (word-level repetition) or if a word is longer than 30 characters (subword-level repetition). 

The results in Table \ref{results_isixhosa_nlg} shows that T-SSLM outperforms tokenisation-based models across reference-based metrics, with substantial BLEU gains. It overcomes text degeneration to a considerable degree for in-language (isiXhosa) pretraining and finetuning, producing much less repetition (Table \ref{table_degen_examples} in the appendix shows examples of generated text). Text degeneration is a challenging problem in low-resource tasks, as evidenced by the high proportions of degenerative examples produced by baselines. 
T-SSLM does not match the performance of massively multilingual LMs \citep{meyer-etal-2024-nglueni}, but our aim is to isolate the effect of task-adaptable subwords over fixed tokenisation. We therefore focus on comparing SSLM to tokenisation-based LMs of the same size trained on the same datasets. In this comparison our results demonstrate the benefit of learnable subwords in data-scarce settings.

We observe similar cross-lingual results, where T-SSLMs pretrained on Setswana/English are finetuned on isiXhosa text generation. T-SSLM consistently outperforms baselines across BLEU and chrF, with competitive text degeneration scores (especially considering reference-based performance differences). This demonstrates the model's ability to improve performance by adapting its subword boundaries to the complex morphology of isiXhosa. These results highlight the potential of learnable subwords, not only for in-language finetuning, but also for cross-lingual transfer in low-resource tasks.

\section{Conclusion}

We investigated the learning dynamics of LMs where the subword segmentation is learnt jointly with knowledge of other aspects of language, in contrast to previous studies that only considered a setup where the subwords segmentation is fixed. 
We identified contrasting patterns in the subword dynamics of Setswana, English, and isi\-Xhosa. 
While learning subword boundaries dynamically may be computationally prohibitive for larger-scale pretraining, 
it enables a new type of LM analysis, providing insight into the role of subwords in different stages of training and demonstrating the potential of task-adaptable segmentation. 
Our findings also underscore the 
distinct subword needs of typologically diverse languages, particularly in low-resource and morphologically complex settings.


\section{Limitations}


\subsection{Computational complexity}  \label{computational_complexity_subsection}

Like other models that marginalise over multiple tokenisations \citep{kreutzer-sokolov-2018-learning, he-etal-2020-dynamic, meyer-buys-2022-subword}, T-SSLM is computationally more expensive than standard LM training. Despite the dynamic program of Eq.~\ref{dp}, marginalisation adds unavoidable cost by computing probabilities for multiple candidate subword sequences (standard LMs compute the likelihood for a single subword sequence). In practice, T-SSLM training times increase by an order of magnitude ($\times10$) over tokenisation-based LMs. 

The added computational complexity of learning subwords prevents us from pretraining on larger datasets. It is an open question whether our findings would scale -- whether similar subword learning trajectories would be observed for larger-scale pretraining and higher-resourced languages. This would require pretraining a large-scale SSLM from scratch on a massive corpus, which is not possible given the compute resources at our disposal. The claims of this paper are limited to smaller-scale pretraining of low-resource languages. This is a relevant and important use case, since data scarcity and computational constraints often co-occur in practice \citep{ahia-etal-2021-low-resource}.

\subsection{Linguistic scope}  

Our findings are limited to three languages, Setswana, English, and isi\-Xhosa, so we cannot guarantee that they will generalise to other language families. We chose these languages based on their morpho-orthopraphic properties, since it allows us to compare the subword learning dynamics of languages with contrasting linguistic subword structures. This contrast simplifies the identification and analysis of differences in subword learning trajectories, making these languages ideal for an initial exploration of subword dynamics.

Our downstream finetuning experiments are limited to isiXhosa data-to-text and headline generation. Given the under-resourcedness of isiXhosa, these are the only text generation evaluation dataset available with sufficient training instances for finetuning. We cannot guarantee similar performance gains for other text generation tasks. Throughout our paper, we are careful not to overstate the impact of our findings in terms of downstream task performance. Instead of positioning T-SSLM as the solution to low-resource text generation, we place more emphasis on the learning dynamics analysis and our finding that pretraining and downstream tasks have different requirements with regards to subword modelling. 

\bibliography{custom, anthology_0, anthology_1}

\appendix

\begin{figure*}[t]
    \begin{center}
        \includegraphics[width=13cm]{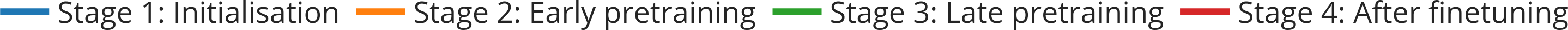}
        \vspace{-0.3cm}
    \end{center}
    
    \subfloat[Setswana]{\includegraphics[width=5.2cm]{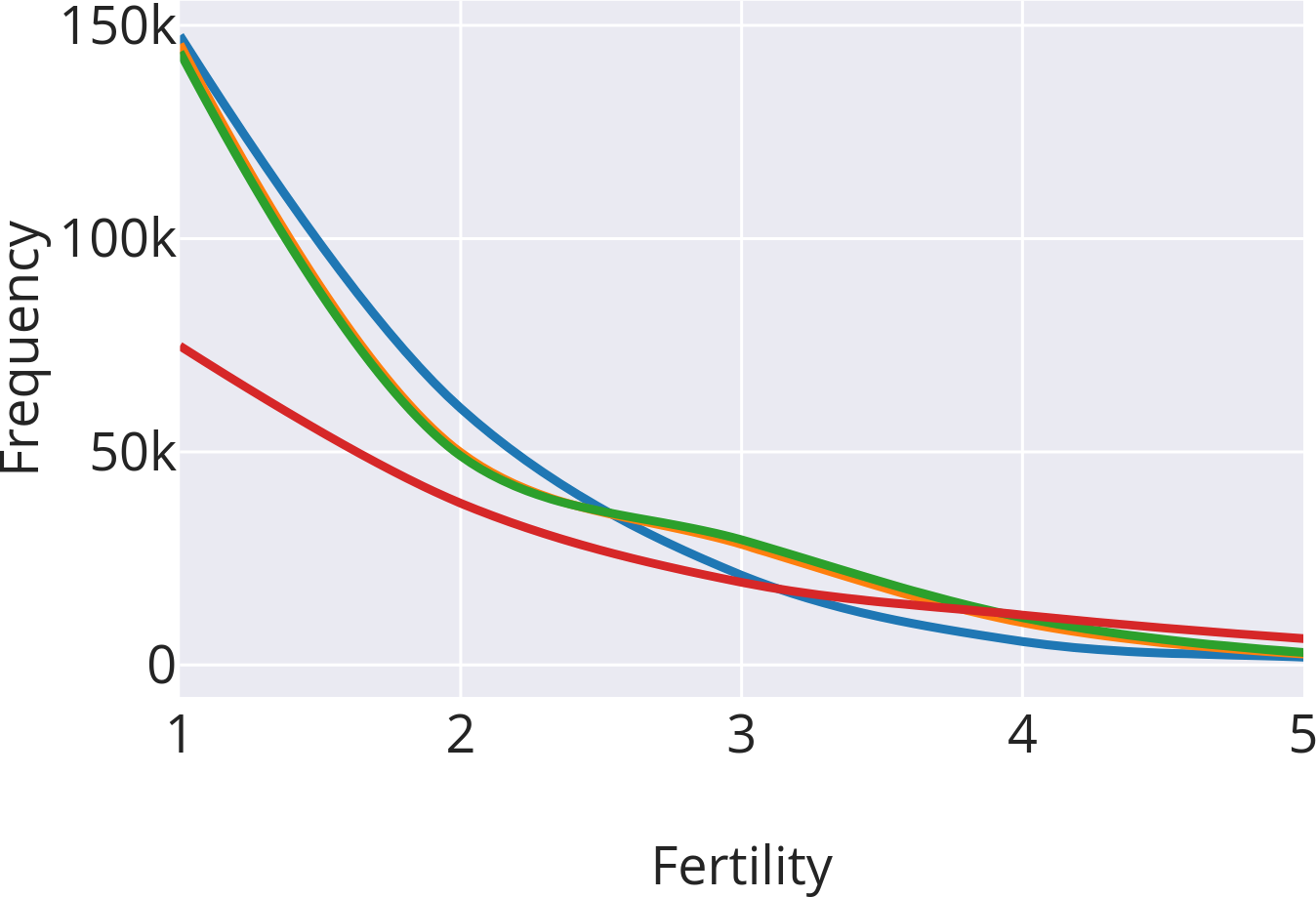}\label{fig:dist:ts}}%
     \,\,
    \subfloat[English]{\includegraphics[width=5.2cm]{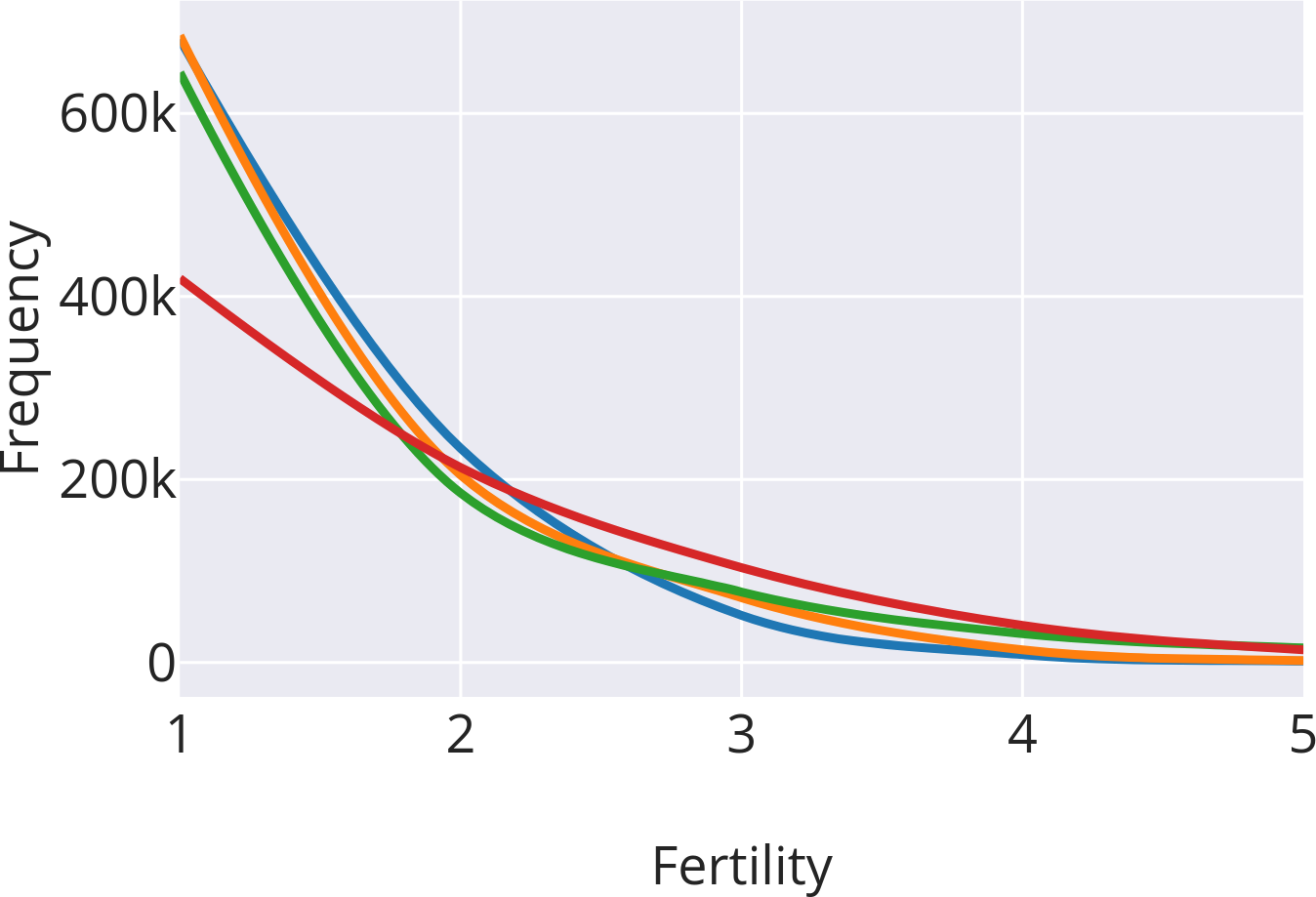}\label{fig:dist:en}}%
    \,\,
    \subfloat[isiXhosa ]{\includegraphics[width=5.2cm]{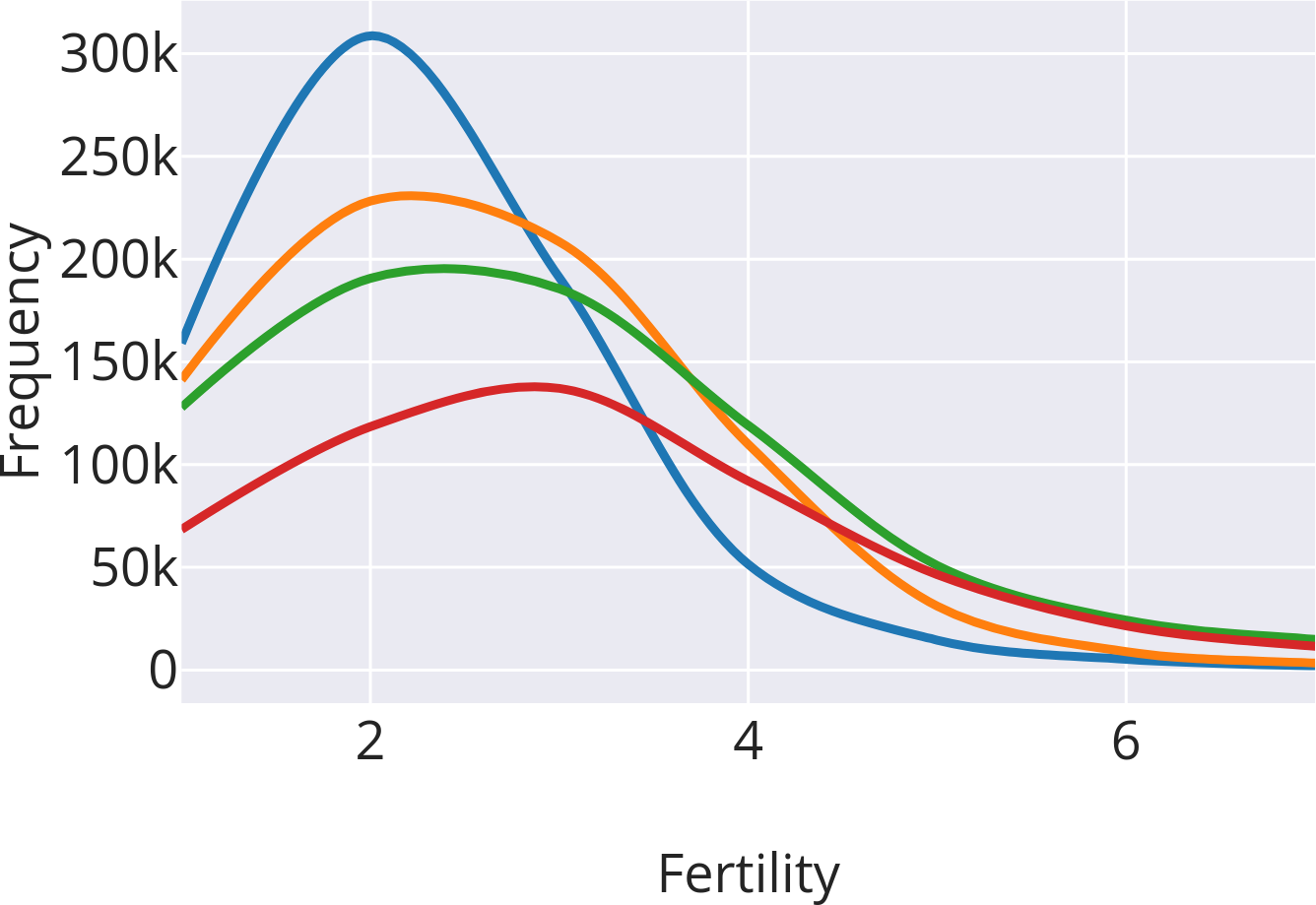}\label{fig:dist:xh}}%
    \caption{Fertility distributions (subwords per segmented word) across the four training stages identified in this study. All languages exhibit increasing fertility as pretraining progresses, with a more pronounced shift during finetuning. Changes are especially dramatic for isiXhosa, whose complex morphology leads to larger distributional shifts than Setswana or English.}%
    \label{fig:dist}%
\end{figure*}

\begin{figure*}[t]%


    \subfloat[Setswana-pretrained]{\includegraphics[width=7.8cm]{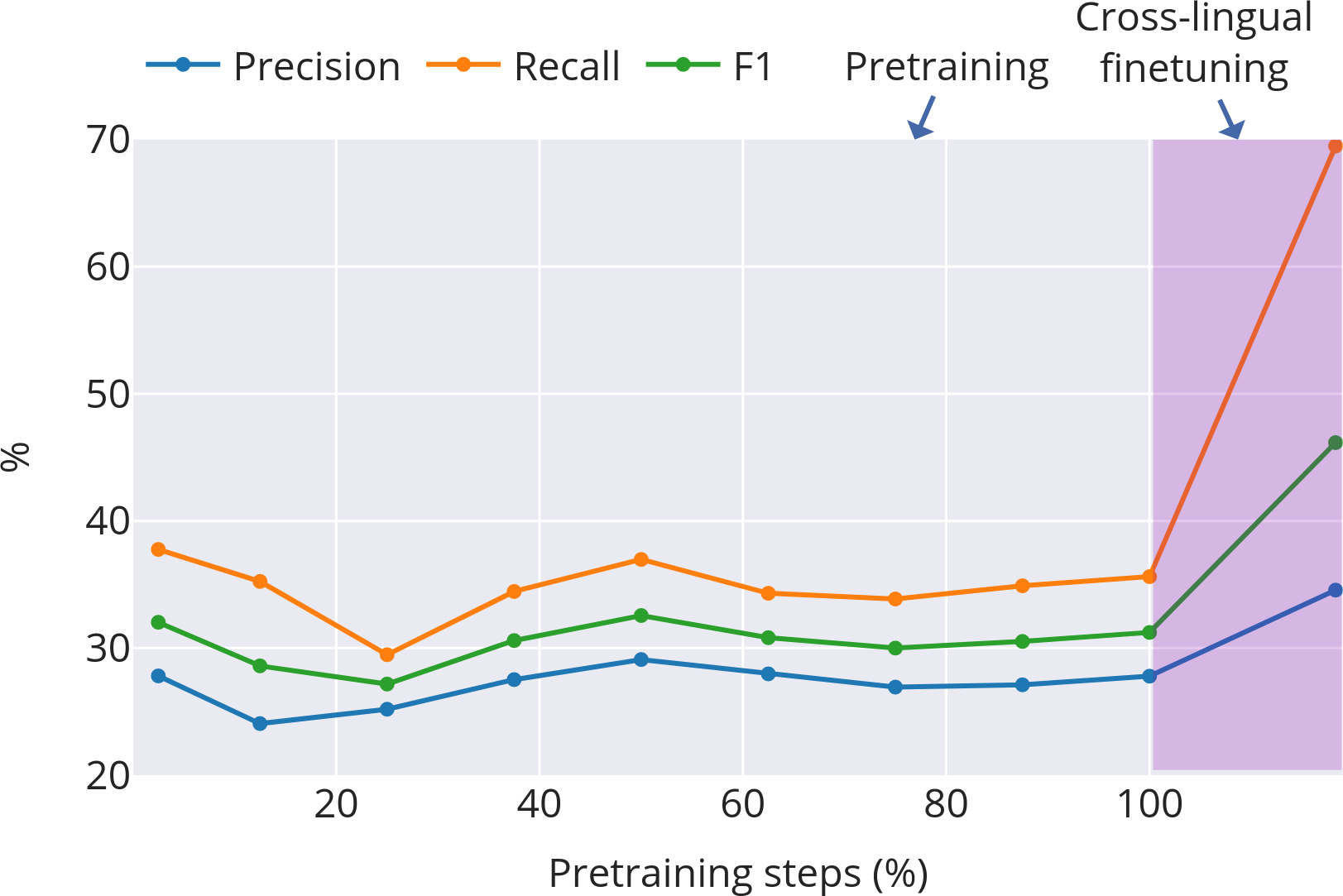}\label{fig:productivity:ts}}%
    \,\,
    \subfloat[English-pretrained]{\includegraphics[width=7.8cm]{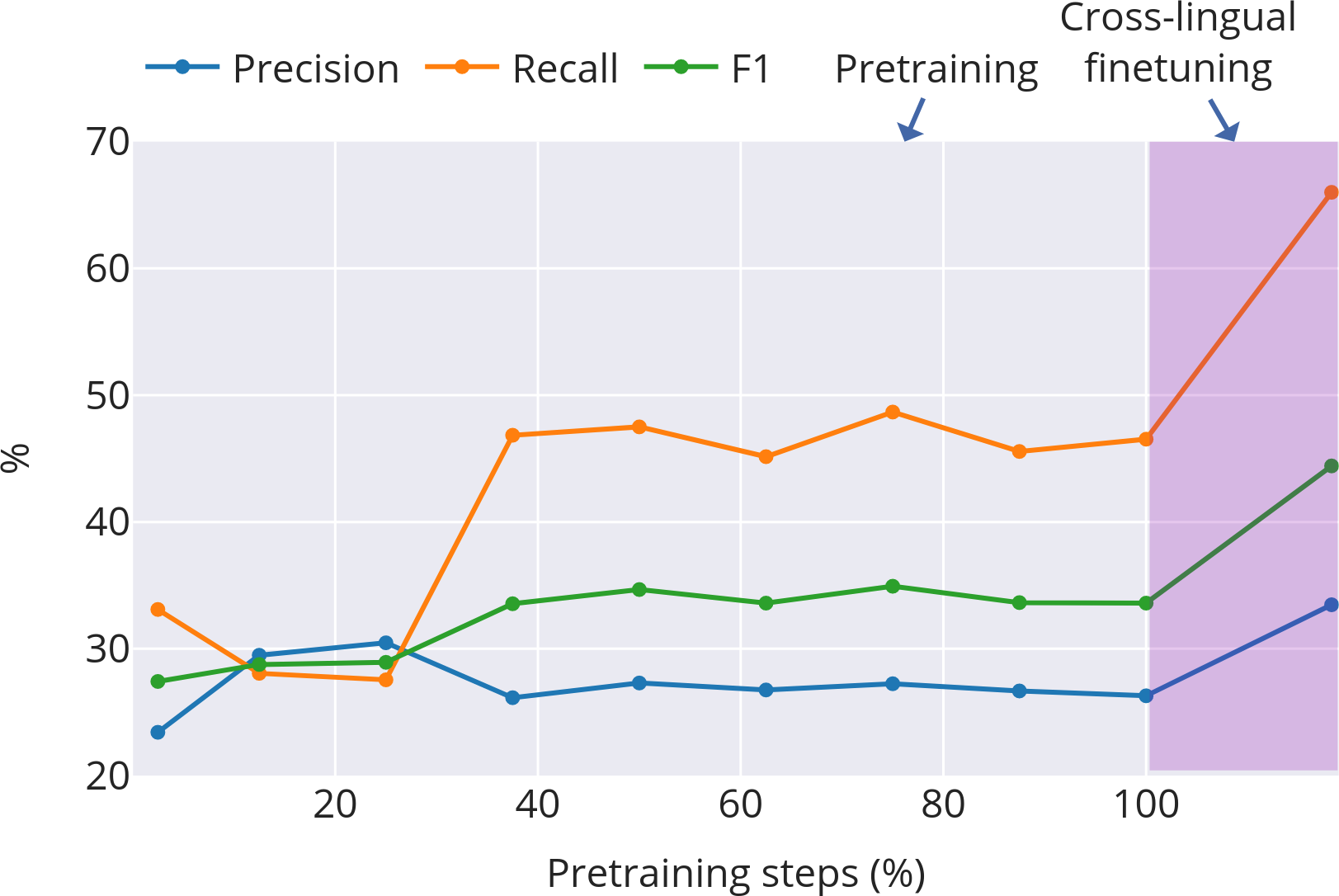}\label{fig:productivity:en}}%

    \caption{Boundary overlap between isiXhosa morphemes and subwords learned by our T-SSLMs pretrained on Setswana (left) and English (right). During isiXhosa finetuning, the models adjusts their subword segmentation to align with the morphological boundaries of the new language.}
    \label{fig:xl-ums}%
\end{figure*}




\begin{table*}[t] 
    \centering
    \begin{tabular}{l} 
    \toprule
         \multicolumn{1}{c}{\textbf{(a) Full prompt template examples for finetuning}} \\
        \midrule        
        South Africa + leaderName + Cyril Ramaphosa \# \# =uCyril Ramaphosa yinkokheli yomzantsi Afrika. \\
        Ethiopia + currency + Ethiopian birr \# \# =Imali yase-Ethiopia yi-Ethiopian Birr. \\
        Denmark + capital + Copenhagen \# \# =ICopenhagen likomkhulu laseDenmark. \\
        \midrule
        \multicolumn{1}{c}{\textbf{(b) Incomplete prompt template examples for testing}}  \\
        \midrule
        Romania + capital + Bucharest \# \# = \\
        India + currency + Indian rupee \# \# = \\
        Netherlands + leaderName + Mark Rutte \# \# = \\
        \bottomrule
    \end{tabular}
    \captionof{table}{To finetune decoder-only PLMS for T2X isiXhosa data-to-text generation, we cast the task as prompt completion. Here we demonstrate the prompt template, ``\texttt{\{input text\} \# \# =\{output text\}}'', with examples from the T2X dataset. During finetuning (a), we only maximise the generation of text after the equals sign ``\texttt{=}''. During testing (b), the special tokens ``\texttt{\# \# =}'' prompt output generation.}
    \label{table_prompts}
\end{table*}

\section{T-SSLM Adaptation} \label{appendix_tsslm}

\subsection{Mixture model} \label{appendix_mixture_subsection}


T-SSLM encodes the preceding text as an untokenised character sequence, using a character-level Transformer. This discards information about segmentation history, but achieves computationally feasible conditioning. For next-subword prediction, we adapt the approach of \citet{meyer-buys-2022-subword} to use a Transformer encoding. 
To condition next-subword prediction on preceding text, we pass the final output embedding to two decoding subnetworks, a character-level model and a softmax layer over a subword lexicon. 

We model next subword probability as a mixture of the two distributions, 
\begin{align} \label{mixture}
    p_{\theta} (t_i | t_{<i}) =\,\, & \phi_i p_{\mathrm{char}} (t_i | t_{<i}) + \nonumber\\
    & (1-\phi_i) p_{\mathrm{lex}} (t_i | t_{<i}),
\end{align}
where $\phi_i$ is a mixture coefficient computed by a fully connected layer $\phi_i = \mathrm{MLP}(t_{<i})$. The subword lexicon contains the $V$ most frequent character $n$-grams in the training corpus. 
The lexicon $p_{\mathrm{lex}}$ models common subwords (e.g. morphemes), while the character-level decoder $p_{\mathrm{char}}$ allows probability assignments to arbitrary character sequences, even those unseen in training (e.g. rare words, names). The lexicon and character components are both softmax layers, jointly trained with the mixture coefficient MLP, which allows T-SSLM to learn in which contexts to rely on the lexicon or character-level predictions.

\subsection{Finetuning algorithm} \label{appendix_finetuning_subsection}

While we can use the existing dynamic programming algorithm of \citet{meyer-buys-2022-subword} for pretraining, we have to adapt it to finetune our T-SSLM for prompt-based text generation. Their dynamic program is represented by Eq. \ref{dp}, which computes (and thereby maximises) the likelihood for a full text sequence. This is not compatible with open-ended text generation, which are cast as text completion tasks: given an input prompt $C$ as context, generate the expected output $O$. The model is finetuned to maximise $p(O|C)$ directly, instead of finetuning on generating both $C$ and $O$. Such completion-only finetuning is standard practice in prompt-based generation.


Suppose $D$ is a completed prompt consisting of $(C, O)$. Instead of maximising $p(D)$, we maximise
\begin{align} \label{finetuning_objective}
    p(O | C) = \frac{p(C, O)}{p(C)} = \frac{p(D)}{p(C)} =\frac{ \alpha_{|C|+|O|}}{\alpha_{|C|}}, 
\end{align}
where $\alpha_{|C|}$ is the forward score up to the end of the input context. 
By maximising this finetuning objective, T-SSLM adapts its subword segmentation to optimise the generation of completion $O$ given input context $C$. This allows the model to adjust the subword boundaries learned during pretraining to optimise performance on the downstream task.

\subsection{Decoding}  \label{appendix_decoding_subsection}

Standard beam search operates on one vocabulary, but the mixture model of T-SSLM combines two vocabularies, subwords and characters, complicating text generation. 
We adapt dynamic decoding, previously proposed for encoder-decoder subword segmental models, for decoder-only generation. 
Dynamic decoding \citep{meyer-buys-2023-subword} generates the target translation one character at a time and predicts subword boundaries based on mixture model probabilities. We adapt this algorithm for our model by conditioning next-character probabilities on preceding prompt context, as opposed to a source-language sentence. 
During generation, we use a beam size of 5 in our dynamic decoding algorithm.

\section{T-SSLM Hyperparameters} \label{appendix_hyperparameters}

\subsection{Pretraining} \label{appendix_sslm_hyperparams}

Our SSLM architecture settings are based on the fairseq \texttt{base\_lm\_architecture}: a 6-layer decoder-only model with 8 attention heads per block, and 512-dimensional embeddings. We pretrain our SSLMs until validation loss stops improving, which occurs by 40 epochs for English and isiXhosa and by 100 epochs for Setswana. 

For isiXhosa pretraining, we use a learning rate of 5e-4 with an inverse square-root scheduler, 4000 warmup steps, and a dropout rate of 0.1. We use an effective batch size of 256 sequences with a maximum sequence length of 512 characters. Our subword lexicon consists of the 10k most frequent character n-grams in the pretraining corpus and we set the maximum subword segment length, a hyperparamter of the SSLM training algorithm, to 5 characters. 
For English and Setswana pretraining, we use the same architecture configuration and training hyperparameters as isiXhosa pretraining, except that we use a subword vocabulary of 5k subwords (since the English and Setswana pretraining corpus are both smaller than the isiXhosa corpus). 

\subsection{Finetuning}

To finetune our LMs, we transform examples into the following template:
\[
\texttt{\{input text\} \# \# = \{output text\}}
\]
During completion-only finetuning (\S\ref{appendix_finetuning_subsection}), we maximise the likelihood of the text following ``\texttt{=}''. Table \ref{table_prompts} contains examples of prompts created for the T2X dataset. For finetuning, we perform a grid search over varying hyperparameter settings and select the final settings based on downstream validation performance. We use a batch size of 16 and a learning rate of 1e-4. 
The learning rate undergoes 500 warmup steps and an inverse square-root scheduler. We perform finetuning for 20 epochs and selected the final model checkpoint based on validation performance.

\section{Baseline Hyperparameters} \label{appendix_baseline_hyperparams}

For BPE, ULM, and BPE-dropout, we use a subword vocabulary size match the subword lexicon of our SSLM (5k for Setswana and English, 10k for isiXhosa). We implement our baselines with the Huggingface Transformers library \citep{wolf-etal-2020-transformers}, training LMs of the same size as our isiXhosa SSLMs (6 layers, 8 attention heads, 512-dimensional embeddings, maximum sequence length of 512). We pretrain the baseline LMs using a batch size of 64 and the default Huggingface pretraining hyperparameters for causal language modelling: a learning rate of 5e-5, no warmup, a linear scheduler, and a dropout rate of 0.1.

To finetune our baselines, we also performed a validation-based grid search over hyperparameter values, settling on a batch size of 4, a learning rate of 5e-5 with no warmup and a linear scheduler. As for our SSLM, we finetuned our baselines for 20 epochs and evaluated the checkpoints with the best validation performance. For decoding, we use standard beam search with beam size of 5.

\begin{table*}[t] 
    \centering
    \begin{tabular}{p{0.17\textwidth}|p{0.775\textwidth}} 
    \toprule

\multicolumn{2}{c}{\textbf{Example (a)}} \\
\midrule
Reference output & UMassimo Drago udlalela uA.S.D. S.S. Nola 1925. \\
\midrule
BPE & UMassimo Drago udlalele i-AS.D.S.\\
\midrule
ULM & Umdlali we-A.S. Nolandelilungu le-A.S. Nolandelilungu le-A.S. Nolandelilungu le-A.S. Nolandelilungu le-A.\\
\midrule
BPE-dropout & UMassimo Drago wadlalela i-A.S. Nola 1925. S. Nola 1925. S. Nola 1925. ... ...\\
\midrule
Char & UMassimo Drago wayengumthetheli kwi - A. S. D. S. S. D. S. S. D. S. D. S. ...\\
\midrule
Byte & UMassimo Drago wayengumthetheli kwi-A.S.D. S.D.S. Nola 1925.\\
\midrule
T-SSLM & UMassimo Drago udlalela i-A.S.D. S.D.S. Nola 1925.\\
\midrule

\multicolumn{2}{c}{\textbf{Example (b)}} \\
\midrule
Reference output & I-St. Vincent-St. Mary school samabanga aphakamileyo siseMelika. \\
\midrule
BPE & ISikhumbuzo i-St. Vincentlanticellanticellanticellanticellanticellanticellanticella...\\
\midrule
ULM & Isikolo seMary High School sise-United States.\\
\midrule
BPE-dropout & ISithili i-St Vincent St. Vincent St.  St. Vincent St. Vincent St. ...\\
\midrule
Char & I - St. Vincentacccccccccccccccccccccccccccccccccccccccccccccccccccccccccc...\\
\midrule
Byte & ISt. Vincentury High School iseUnited States.\\
\midrule
T-SSLM & St. Vincent-St. Mary High School ifumaneka e United States.\\
        \bottomrule
    \end{tabular}
    \captionof{table}{Examples of model generations on the T2X test set, compared to reference output texts. The tokenisation-based baselines often generate incoherent, repetitive text, while SSLM avoids text degeneration in all instances.}
    \label{table_degen_examples}
\end{table*}

\end{document}